\documentclass[journal, twocolumns]{IEEEtran}

\usepackage{polpack}
\usepackage{soul} 
\usepackage{graphics}
\usepackage{color}
\usepackage[usenames]{xcolor}

\begin{document}
\title{A Learning Theoretic Approach to Energy Harvesting Communication System Optimization}
\author{\IEEEauthorblockN{Pol Blasco\IEEEauthorrefmark{1}},  \IEEEauthorblockN{Deniz G\"{u}nd\"{u}z\IEEEauthorrefmark{2}} and \IEEEauthorblockN{Mischa Dohler\IEEEauthorrefmark{1}}
\\
\IEEEauthorblockA{\IEEEauthorrefmark{1}
CTTC, Barcelona, Spain
\\
Emails:\{pol.blasco, mischa.dohler\}@cttc.es}
\\
\IEEEauthorblockA{\IEEEauthorrefmark{2}
Imperial College London, United Kingdom
\\
Email: d.gunduz@imperial.ac.uk }

\thanks{This work was partially supported by the European Commission in the framework of EXALTED-258512, ACROPOLIS NoE ICT-2009.1.1 and Marie Curie IRG Fellowship with reference number 256410 (COOPMEDIA) and by the Spanish Government under SOFOCLES Grant TEC2010-21100, FPU Grant with reference AP2009-5009 and  TEC2010-17816 (JUNTOS).}
}

\maketitle

\begin{abstract}
A point-to-point wireless communication system in which the transmitter is equipped with an  energy harvesting device and a rechargeable battery, is studied. Both the energy and the data arrivals at the transmitter are modeled as Markov processes. Delay-limited communication is considered assuming that the underlying channel is block fading with memory, and the instantaneous channel state information is available at both the transmitter and the receiver. The expected total transmitted data  during the transmitter's activation time is maximized under three different sets of assumptions regarding the information available at the transmitter about the underlying stochastic processes. A \emph{learning theoretic approach} is introduced, which does not assume any \emph{a priori} information on the Markov processes governing the communication system. In addition, \emph{online} and \emph{offline} optimization problems are studied for the same setting. Full statistical knowledge and causal information on the realizations of the underlying stochastic processes are assumed in the online optimization problem, while the offline optimization problem assumes non-causal knowledge of the realizations in advance. Comparing the optimal solutions in all three frameworks, the performance loss due to the lack of the transmitter's information regarding the behaviors of the underlying Markov processes is quantified.
\end{abstract}

\section{Introduction} \label{sec:intro}
Energy harvesting (EH) has emerged as a promising technology  to extend the lifetime of communication networks, such as machine-to-machine or wireless sensor networks; complementing current battery-powered transceivers by harvesting the available ambient energy (solar, vibration, thermo-gradient, etc.). As opposed to battery limited devices, an EH transmitter can theoretically operate over an unlimited time horizon; however, in practice transmitter's activation time is limited by other factors and typically the harvested energy rates are quite low. Hence, in order to optimize the communication performance, with sporadic arrival of energy in limited amounts, it is critical to optimize the transmission policy using the available information regarding the energy and data arrival processes.

There has been a growing  interest in the optimization of EH communication systems. Prior research can be grouped into two, based on the information (about the energy and data arrival processes) assumed to be available at the transmitter. In the \textit{offline optimization} framework, it is assumed that the transmitter has non-causal information on the exact data/energy arrival instants and amounts~\cite{Yang:TC:10,Devillers:JCN:11,Orhan2012,Tutuncuoglu:JCN:12,Antepli:JSAC:11,Huang:JSAC:12,Gunduz:CAMSAP:11,Ozel:JSAC:11,hc:gregori2012}. In the \textit{online optimization} framework, the transmitter is assumed to know the statistics of the underlying EH and data arrival processes; and has causal information about their realizations~\cite{hj:Lei2009,hc:Sinha2011,hj:Wang2012,hj:Li2011,eh:HoZhang11,hj:Srivastava2011,hj:Mao2012}.

Nonetheless, in many practical scenarios either the characteristics of the EH and data arrival processes change over time, or it is not possible to have reliable statistical information about these processes before deploying the transmitters. For example, in a sensor network with solar EH nodes distributed randomly over a geographical area, the characteristics of each node's harvested energy will depend  on its location, and will change based on the time of the day or the season. Moreover, non-causal information about the data/energy arrival instants and amounts is too optimistic in practice, unless the underlying EH process is highly deterministic. Hence, neither online nor offline optimization frameworks will be satisfactory in most practical scenarios. To adapt the transmission scheme to the unknown EH and data arrival processes, we propose a \emph{learning theoretic approach}.

We consider a point-to-point wireless communication system in which the transmitter is equipped with an EH device and a finite-capacity rechargeable battery. Data and energy arrive at the transmitter in packets in a time-slotted fashion. At the beginning of each time-slot (\TS), a data packet arrives and it is lost if not transmitted within the following \TS. This can be either due to the strict delay requirement of the underlying application, or due to the lack of a data buffer at the transmitter.  Harvested energy can be stored in a finite size battery/capacitor for future use, and we consider that the transmission of data is the only source of energy consumption. We assume that the wireless channel between the transmitter and the receiver is constant for the duration of a \TS but may vary from one \TS to the next. We model the data and energy packet arrivals as well as the channel state as Markov processes. The activation time of an EH transmitter is not limited by the available energy; however, to be more realistic we assume that the transmitter might terminate its operation at any \TS with certain probability. This can be due to physical limitations, such as  blockage of its channel to the receiver, failure of its components, or because it is forced to switch to the idle mode by the network controller. The objective of the transmitter is to maximize the expected total transmitted data to the destination during its activation time under the  energy  availability constraint and the individual deadline constraint for each packet.

For this setting, we provide a complete analysis of the optimal system operation studying the offline, online and the learning theoretic optimization problems. The solution for the offline optimization problem constitutes an upperbound on the online optimization, and the difference between the two indicates the value of knowing the system behavior non-causally.
In the learning-based optimization problem we take a more practically relevant approach, and assume that the statistical information about the underlying Markov processes  is not available at the transmitter, and that, all the data and energy arrivals as well as the channel states are known only causally. Under these assumptions, we propose a machine learning algorithm for the transmitter operation, such that the transmitter learns the optimal transmission policy over time by performing actions and observing their immediate rewards. We show that the performance of the proposed learning algorithm converges to the solution of the online optimization problem as the learning time increases. The main technical contributions of the paper are summarized as follows:
\begin{itemize}
  \item We provide, to the best of our knowledge, the first learning theoretic optimization approach to the EH communication system optimization problem under stochastic data and energy arrivals.
  \item For the same system model, we provide a complete analysis by finding the optimal transmission policy for both the online and offline optimization approaches in addition to the learning theoretic approach.
  \item For the learning theoretic problem, we propose a \mbox{Q-learning} algorithm and show that its performance converges to that of the optimal online transmission policy as the learning time increases.
  \item For the online optimization problem, we propose and analyze a transmission strategy based on the policy iteration algorithm.
  \item We show that the offline optimization problem can be written as a mixed integer linear program. We provide a solution to this problem through the branch-and-bound algorithm. We also propose and solve a linear program relaxation of the offline optimization problem.
  \item We provide a number of numerical results to corroborate our findings, and compare the performance of the learning theoretic optimization with the offline and online optimization solutions numerically.
\end{itemize}

The rest of this paper is organized as follows. Section~\ref{sec:rw} is dedicated to a summary of the related literature. In Section~\ref{sec:system}, we present the EH communication system model. In Section~\ref{sec:online}, we study the online optimization problem and characterize the optimal transmission policy. In Section~\ref{sec:learning}, we propose a learning theoretic approach, and show that the transmitter is able to learn the stochastic system dynamics and converge to the optimal transmission policy.  The offline optimization problem is studied in Section~\ref{sec:offline}. Finally in Section~\ref{sec:results}, the three approaches are compared and contrasted in different settings through numerical analysis. Section~\ref{sec:conclusions} concludes the paper.

\section{Related Work}\label{sec:rw}

There is a growing literature on the optimization of EH communication system within both online and offline optimization frameworks. Optimal offline transmission strategies have been characterized for point-to-point systems with both data and energy arrivals in~\cite{Yang:TC:10}, with battery imperfections in~\cite{Devillers:JCN:11}, and with processing energy cost in~\cite{Orhan2012}; for various multi-user scenarios in~\cite{Devillers:JCN:11,Tutuncuoglu:JCN:12,Antepli:JSAC:11,Huang:JSAC:12,Gunduz:CAMSAP:11}; and for fading channels in~\cite{Ozel:JSAC:11}. Offline optimization of precoding strategies for a MIMO channel is studied in~\cite{hc:gregori2012}. In the online framework the system is modeled as a Markov decision process (MDP) and dynamic programming (DP)~\cite{dp:Bellman1957a} based solutions are provided. In~\cite{hj:Lei2009}, the authors assume that the packets arrive as a Poisson process, and each packet has an intrinsic value assigned to it, which also is a random variable. Modeling the battery state as a Markov process, the authors study the optimal transmission policy that maximizes the average value of the received packets at the destination. Under a similar Markov model, \cite{hc:Sinha2011} studies the properties of the optimal transmission policy. In~\cite{hj:Wang2012}, the minimum transmission error problem is addressed, where the data and energy arrivals are modeled as Bernoulli and  Markov processes, respectively. Ozel et al. \cite{Ozel:JSAC:11} study online as well as offline optimization of a throughput maximization problem with stochastic energy arrivals and a fading channel. The causal information assumption is relaxed by modeling the system as a partially observable MDP in~\cite{hj:Li2011} and~\cite{eh:HoZhang11}. Assuming that the data and energy arrival rates are known at the transmitter, tools from queueing theory are used for long-term average rate optimization in~\cite{hj:Srivastava2011} and~\cite{hj:Mao2012} for point-to-point and multi-hop scenarios, respectively.

Similar to the present paper, references~\cite{Kansal2003,hc:Hsu2006,hc:Vigorito2007,jh:Hsu2009} optimize EH communication systems under mild assumptions regarding the  statistical information available at the transmitter. In~\cite{Kansal2003} a forecast method for a periodic EH process is considered. Reference~\cite{hc:Hsu2006} uses historical data to forecast energy arrival and solves a duty cycle optimization problem based on the expected energy arrival profile. Similarly to~\cite{hc:Hsu2006},  the  transmitter duty cycle is optimized in \cite{hc:Vigorito2007} and~\cite{jh:Hsu2009} by taking advantage of techniques from control theory and machine learning, respectively.  However, \cite{hc:Hsu2006,hc:Vigorito2007,jh:Hsu2009} consider only balancing  the harvested and consumed energy regardless of the underlying data arrival process and the cost associated to data transmission. In contrast, in our problem setting we consider the data arrival and channel state processes together with the EH process, significantly complicating the problem.

\section{System Model}\label{sec:system}
We consider a wireless transmitter equipped with an EH device and a rechargeable battery with limited storage capacity. The communication system operates in a time-slotted fashion over \TSD of equal duration. We assume that both data and energy arrive in packets at each \TS. The channel state remains constant during each \TS and changes from one \TS to the next. We consider strict delay constraints for the transmission of data packets; that is, each data packet needs to be transmitted within the \TS following its arrival. We assume that the transmitter has a certain small probability ($1-\gamma$) of terminating its operation at each \TS, and it is interested in maximizing the expected total transmitted data during its activation time.

The sizes of the data/energy packets arriving at the beginning of each \TS are modeled as correlated time processes following a first-order discrete-time Markov model. Let \Ds be the size of the data packet arriving at \TSn, where $\Ds \in \Dset \triangleq \{\Delement[1],\dots,\Delement[\ND]\}$, and \ND is the number of elements in \Dset. Let \Dtr{\Delement[k]} be the probability of the data packet size process going from state \Delement to state \Delement[k] in one \TS. Each energy packet is assumed to be an integer multiple of a fundamental energy unit. Let \Ehs denote the amount of energy harvested during \TSn, where $\Ehs\in\Ehset\triangleq\{\Ehelement[1],\dots,\Ehelement[\NE]\}$, and \Ehtr{\Ehelement[k]} is the state transition probability function. The energy harvested during \TSn, \Ehs, is stored in the battery and can be used for data transmission at the beginning of \TSn[\tindex+1]. The battery has a limited size of \NB energy units and all the energy harvested when the battery is full is lost. Let \Cs be the channel state during \TSn, where $\Cs\in\Cset\triangleq\{\Celement[1],\dots,\Celement[\NC]\}$. We assume that \Cs also follows a Markov model; \Ctr{\Celement[k]} denotes its state transition probability, and the realization of \Cs at each \TSn is known at the receiver. Similar models have been considered for EH~\cite{eh:HoZhang11,hj:Li2011,hj:Wang2012}, data arrival~\cite{hj:Li2011}, and channel state~\cite{hc:Aprem2012,eh:HoZhang11} processes. Similar to our model, \cite{hj:Lei2009} also considers a strict deadline constraint and lack of data buffer at the transmitter.

For each channel state \Cs and  packet size \Ds, the transmitter knows the amount of minimum energy \Emin required to transmit the arriving data packet to the destination. Let $\Emin=\fDE{\Cs}:\Dset\times\Cset\rightarrow\Eunit$ where  \Eunit is a discrete set of integer multiples of the fundamental energy unit. We assume that if the transmitter spends \Emin units of energy the packet is transmitted successfully.

\begin{figure}
  \centering
  \includegraphics[width=0.5\textwidth]{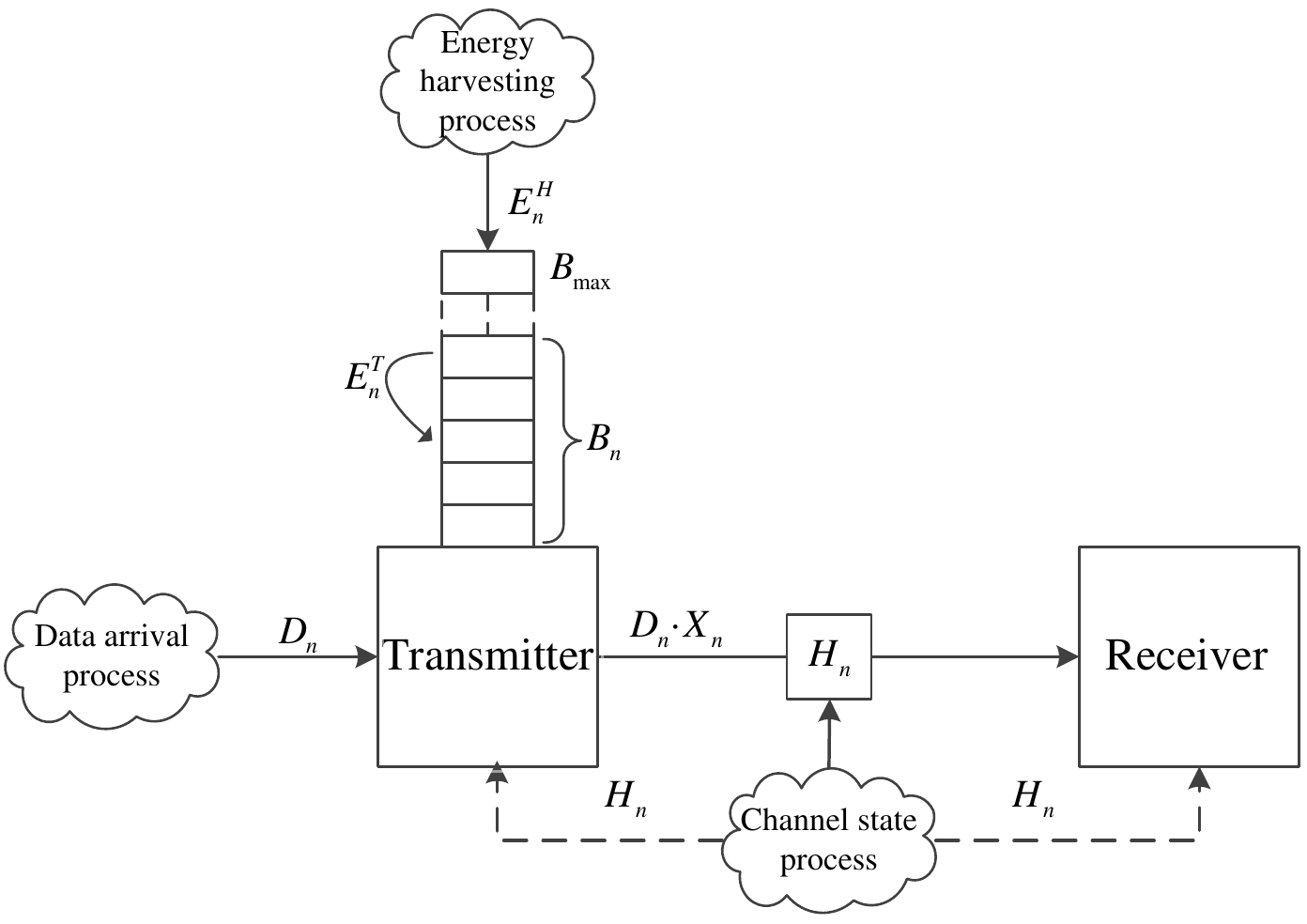}\\ \vspace{-0.5cm}
  \caption{EH communication system with EH and data arrival stochastic processes as well as varying channel.}\label{fig:sysmodel}
\end{figure}

In each \TSn, the transmitter knows the battery state \Bs, the size of the arriving packet \Ds, the current channel state \Cs; and hence, the amount of energy \Emin required to transmit this packet. At the beginning of each \TS, the transmitter makes a binary decision: to transmit or to drop the incoming packet. This may account for the case of control or measurement packets, where the data in the packet is meaningful only if received as a whole. Additionally, the transmission rate and power are fixed at the beginning of each \TS, and cannot be changed within the \TS. The transmitter must guarantee that the energy spent in \TSn is not greater than the energy available in the battery, \Bs. Let $\xt\in\{0,1\}$ be the indicator function of the event that the incoming packet in \TSn is transmitted. Then, for  $\forall \tindex \in \mathds{Z}$, we have
\begin{align}
\xt\Emin &\leq \Bs,\label{eq:battery_1}\\
\Bs[\tindex+1]&= \min \{\Bs-\xt \Emin+\Ehs, \NB \}.\label{eq:battery_2}
\end{align}
The goal is to maximize the expected total transmitted data over the activation time of the transmitter, which is given by:
\begin{equation} \label{eq:opt_problemv1}
\begin{aligned}
\max_{\{\xt[i]\}_{i=0}^{\infty}} & \lim_{\NTS \rightarrow \infty} \expected{\sum_{\tindex=0}^{\NTS} \gamma^\tindex\xt \Ds}{},\\
\text{s.t. } & (\ref{eq:battery_1}) \text{ and } (\ref{eq:battery_2}),
\end{aligned}
\end{equation}
where $0 < 1-\gamma \leq 1$ is the independent and identically distributed probability of the transmitter to terminate its operation in each \TS. We call this problem the \emph{expected total transmitted data maximization problem} (\DP) as the transmitter aims at maximizing the total transmitted data during an unknown activation time. The EH system that is considered here is depicted in Figure~\ref{fig:sysmodel}.

 We will also consider the case with $\gamma=1$; that is, the transmitter can continue its operation as long as there is available energy. In this case, contrary to the \DP, (\ref{eq:opt_problemv1}) is not a practical measure of performance as the transmitter operates for an infinite amount of time;  and hence, most transmission policies that allow a certain non-zero probability of transmission at each \TS are optimal in the expected total transmitted data criterion as they all transmit an infinite amount of data. Hence, we focus on the \emph{expected throughput maximization problem} (\TP):
\begin{equation} \label{eq:opt_problemv3}
\begin{aligned}
\max_{\{\xt[i]\}_{i=0}^{\infty}} & \lim_{\NTS \rightarrow \infty} \frac{1}{N+1}\expected{\sum_{\tindex=0}^{\NTS} \xt \Ds}{},\\
\text{s.t. } & (\ref{eq:battery_1}) \text{ and } (\ref{eq:battery_2}).
\end{aligned}
\end{equation}
The main focus of the paper is on the \DP, therefore, we assume $0\leq\gamma<1$ in the rest of the paper unless otherwise stated. The \TP will be studied numerically in Section~\ref{sec:results}.

An MDP provides a mathematical framework for modeling decision-making situations where outcomes are partly random and partly under the control of the decision maker \cite{dp:Bellman1957b}.  The EH communication system, as described above, constitutes a finite-state discrete-time MDP. An MDP is  defined via the quadruplet $\langle \Wset,\mathcal{A}, \Wtr{\Welement[k]}{x_i}, \WR{\Welement[k]}{x_i}\rangle$, where \Wset is the set of possible states, $\mathcal{A}$ is the set of actions, \Wtr{\Welement[k]}{x_i} denotes the transition probability from state \Welement to state \Welement[k] when action $x_i$ is taken, and \WR{\Welement[k]}{x_i} is the immediate reward yielded when in state \Welement action $x_i$ is taken and the state changes to \Welement[k]. In our model the state of the system in \TSn is \Ws, which is formed by four components $\Ws = (\Ehs,\Ds,\Cs,\Bs)$. Since all components of \Ws are discrete there exist a finite number of possible states and the set of states is denoted by $\Wset=\{\Welement[1],\dots,\Welement[\NW]\}$. The set of actions is $\mathcal{A} = \{0, 1\}$, where action $0$ ($1$) indicate that the packet is dropped (transmitted). If the immediate reward yielded by action $x_i \in \mathcal{A}$ when the state changes from \Ws to \Ws[\tindex+1] in \TSn is $\WR[\Ws]{\Ws[\tindex+1]}{x_i}$, the objective of an MDP is to find the optimal transmission policy $\pi(\cdot):\Wset\rightarrow\mathcal{A}$ that maximizes the expected discounted sum reward (i.e., the expected total transmitted data). We restrict our attention to deterministic stationary transmission policies. In our problem, the immediate reward function is $\WR[\Ws]{\Ws[\tindex+1]}{\xt} = \xt\Ds$, and the expected discounted sum reward is equivalent to (\ref{eq:opt_problemv1}), where $\gamma$ corresponds to the discount factor, and $\xt=\pi(\Ws)$ is the action taken by the transmitter when the system is in state \Ws. 

Given the policy $\pi$ and the current state \Ws, the state of the battery \Bs[\tindex+1] is ubiquitously  determined by (\ref{eq:battery_2}). The other state components are randomly determined using the state transition probability functions. Since state transitions depend only on the current state and the transmitter's current action, the model under consideration fulfills the Markov property. As a consequence, we can take advantage of DP and reinforcement learning (RL)~\cite{rl:sutton&barto} tools to solve the \DP.

Next, we introduce the \emph{state-value function} and \emph{action-value function} which will be instrumental in solving the MDP \cite{rl:sutton&barto}. The state-value function is defined as follows:
\begin{equation}\label{eq:Vfun}
V^\pi(\Welement)\triangleq\! \sum_{\forall \Welement[k]\in \Wset}\Wtr{\Welement[k]}{\pi(\Welement)}\left[\WR{\Welement[k]}{\pi(\Welement)} +\gamma V^{\pi}(\Welement[k]) \right]\!.\!
\end{equation}
It is, intuitively, the expected discounted sum reward of policy $\pi$ when the system is in state \Welement. The action-value function, defined as
 \begin{equation}\label{eq:Qfun}
Q^\pi(\Welement,x_i)\triangleq \sum_{\forall \Welement[k]\in \Wset}\Wtr{\Welement[k]}{x_i}\left[\WR{\Welement[k]}{x_i} +\gamma V^{\pi}(\Welement[k]) \right],
\end{equation}
is the expected discounted reward when the system is in state \Welement, takes action $x_i\in \mathcal{A}$, and follows policy $\pi$ thereafter. A policy $\pi$ is said to be better than or equal to policy $\pi'$, denoted by $\pi \geq \pi'$, if the expected discounted reward of $\pi$ is greater than or equal to that of $\pi'$ in all states, i.e., $\pi \geq \pi'$ if $V^\pi(\Welement) \geq V^{\pi'}(\Welement), \forall \Welement \in \Wset$. The optimal policy $\pi^*$ is the policy that is better than or equal to any other policy. Eqn. (\ref{eq:Vfun})  indicates that the state-value function $V^\pi(\Ws)$ can be expressed as a combination of the expected immediate reward and the state value function of the next state, $V^\pi(\Ws[\tindex+1])$. The same happens with the action-value function. The state-value function when the transmitter follows the optimal policy is
\begin{equation} \label{eq:Vopt}
V^{\pi^*}(\Welement)=\max_{x_j\in \mathcal{A}}Q^{\pi^*}(\Welement,x_j).
\end{equation}
From (\ref{eq:Vopt}) we see that the optimal policy is the \emph{greedy policy}; that is, the policy that performs the action with the highest expected discount reward according to $Q^{\pi^*}(\Welement,x_j)$. The action-value function, when the optimal policy is followed, is
\begin{equation} \small \label{eq:Qopt}
Q^{\pi^*}(\Welement,x_i)=\sum_{\forall \Welement[k]\in \Wset}\Wtr{\Welement[k]}{x_i} \bigl [ \WR{\Welement[k]}{x_i} +\gamma \max_{x_j\in \mathcal{A}}Q^{\pi^*}(\Welement[k],x_j) \bigr ].
\end{equation}

Similarly to (\ref{eq:Vfun}),  (\ref{eq:Qopt})  indicates that the action-value function $Q^{\pi^*}(\Ws,x_i)$, when following $\pi^*$, can be expressed as a combination of the expected immediate reward and the maximum value of the action-value function of the next state.

There are three approaches to solve the \DP depending on the available information at the transmitter. If the transmitter has prior information on the values of \Wtr{\Welement[k]}{x_i} and \WR{\Welement[k]}{x_i}, the problem falls into the \emph{online optimization} framework, and we can use DP to find the optimal transmission policy $\pi^*$. If the transmitter does not have prior information on the values of \Wtr{\Welement[k]}{x_i} or \WR{\Welement[k]}{x_i} we can use a \emph{learning theoretic} approach based on RL. By performing actions and observing their rewards, RL  tries to arrive at an optimal policy $\pi^*$ which maximizes the expected discounted sum reward accumulated over time. Alternatively, in the \emph{offline optimization} framework, it is assumed that all future EH states \Ehs, packet sizes \Ds and channel states \Cs are known non-causally over a finite horizon.
\begin{rem}
If the transmitter is allowed to transmit a smaller portion of each packet, using less energy than required to transmit the whole packet, one can re-define the finite action set $\mathcal{A}$. As long as the total number of actions and states remains finite, all the optimization algorithms that we propose in Sections \ref{sec:online} and \ref{sec:learning}  remains to be valid. In principle, DP and RL ideas can be applied to problems with continuous state and action spaces as well; however, exact solutions are possible only in special cases. A common way of obtaining approximate solutions with continuous state and action spaces is to use function approximation techniques~\cite{rl:sutton&barto}; e.g., by discretizing the action space into a finite set of packet portions, or using fuzzy \mbox{Q-learning}~\cite{Glorennec1997}.
\end{rem}

\section{Online Optimization}\label{sec:online}
 We first consider the online optimization problem. We employ \emph{policy iteration} (PI)~\cite{dp:Putterman2005}, a DP algorithm, to find the optimal policy in (\ref{eq:opt_problemv1}). The MDP problem in (\ref{eq:opt_problemv1}) has finite action and state spaces as well as bounded and stationary immediate reward functions. Under these conditions PI is proven to converge to the optimal policy when $0\leq \gamma < 1$~\cite{dp:Putterman2005}. The key idea is to use the structure of (\ref{eq:Vfun}), (\ref{eq:Qfun}) and (\ref{eq:Vopt}) to obtain the optimal policy. PI is based on two steps: 1) policy evaluation, and 2) policy improvement.

In the policy evaluation step the value of a policy $\pi$ is evaluated by computing the value function $V^{\pi}(\Welement)$. In principle, (\ref{eq:Vfun}) is solvable  but at the expense of laborious calculations when \Wset is large. Instead, PI uses an iterative method~\cite{rl:sutton&barto}: given $\pi$, \Wtr{\Welement[k]}{x_i} and  \WR{\Welement[k]}{x_i}, the state value function $V^{\pi}(\Welement)$ is estimated as
\begin{equation} \label{eq:Vest}
\!V_l^{\pi}(\Welement)\!=\!\sum_{\Welement[k]}\Wtr{\Welement[k]}{\pi(\Welement)}\left[\WR{\Welement[k]}{\pi(\Welement)} +\gamma V_{l-1}^{\pi}(\Welement[k]) \right]\!,\!\!\!
\end{equation}
for all $\Welement \in \Wset$, where $l$ is the iteration number of the estimation process. It can be shown that the sequence $V_l^{\pi}(\Welement)$ converges to $V^\pi(\Welement)$ as $l \rightarrow \infty$ when $0\leq \gamma<1$. With policy evaluation, one evaluates how good a policy $\pi$ is by computing its expected discounted reward at each state $\Welement\in\Wset$.

In the policy improvement step, the PI algorithm looks for a policy $\pi'$ that is better than the previously evaluated policy $\pi$. The Policy Improvement Theorem~\cite{dp:Bellman1957a} states that if $Q^{\pi}(\Welement,\pi' (\Welement))\geq V^{\pi}(\Welement)$ for all $\Welement \in \Wset$ then  $\pi' \geq \pi$. Policy improvement step finds the new policy $\pi'$ by applying the greedy policy to $Q^\pi(\Welement,x_i)$ in each state. Accordingly, the new policy $\pi'$ is selected as follows:
\begin{equation}
\pi'(\Welement)=\argmax_{x_i \in \mathcal{A}} Q^{\pi}(\Welement,x_i).
\end{equation}
PI works iteratively by first evaluating $V^{\pi}(\Welement)$, finding a better policy $\pi'$, then evaluating $V^{\pi'}(\Welement)$, and finding a better policy $\pi''$, and so forth. When the same policy is found in two consecutive iterations we conclude that the algorithm has converged. The exact embodiment of the algorithm, as described in~\cite{rl:sutton&barto}, is given in Algorithm \ref{alg:PI}. The worst-case complexity of PI depends on the number of states, $N_{\mathcal{S}}$,  and actions; and in our particular model, the complexity of PI is bounded by $O\left(\frac{2^{N_{\mathcal{S}}}}{N_{\mathcal{S}}}\right)$ \cite{Mansour1999}.  The performance of the proposed algorithm and the comparison with other approaches will be given in Section~\ref{sec:results}.

\begin{algorithm}[t]
\small\caption{Policy Iteration (PI)} \label{alg:PI}
\begin{algorithmic}
   \STATE \textbf{$\mathbf{1.}$ Initialize:}
   \FOR{each $\Welement \in \Wset$}
           \STATE initialize $V(\Welement)$ and $\pi(\Welement)$ arbitrarily
   \ENDFOR
  \STATE \textbf{$\mathbf{2.}$ Policy evaluation:}
   \REPEAT
            \STATE $\Delta \leftarrow 0$
            \FOR{each $\Welement \in \Wset$}
                \STATE $v \leftarrow V(\Welement)$
                \STATE $V(\Welement) \leftarrow \sum_{\Welement[k]}\Wtr{\Welement[k]}{\pi(\Welement)}\left[\WR{\Welement[k]}{\pi(\Welement)} +\gamma V(\Welement[k]) \right]$
                \STATE  $\Delta \leftarrow \max(\Delta, \| v-V(\Welement) \|)$
            \ENDFOR
   \UNTIL $\Delta < \epsilon$
 \STATE \textbf{$\mathbf{3.}$ Policy improvement:}
\STATE policy-stable $\leftarrow$ true
 \FOR{each $\Welement \in \Wset$}
    \STATE $b\leftarrow \pi(\Welement)$
    \STATE $\pi(\Welement)\leftarrow \argmax_{x_i \in \mathcal{A}} \sum_{\Welement[k]}\Wtr{\Welement[k]}{x_i}\left[\WR{\Welement[k]}{x_i} +\gamma V(\Welement[k]) \right]$
    \IF{$b \neq \pi(\Welement)$}
    \STATE policy-stable $\leftarrow$ false
    \ENDIF
  \ENDFOR
 \STATE \textbf{$\mathbf{4.}$ Check stoping criteria:}
\IF{policy-stable}
\STATE stop
\ELSE
\STATE go to $2$).
\ENDIF
\end{algorithmic}
\end{algorithm}

\section{Learning Theoretic Approach}\label{sec:learning}
Next we assume that  the transmitter has no knowledge of the transition probabilities \Wtr{\Welement[k]}{x_i} and the immediate reward function \WR{\Welement[k]}{x_i}. We use \mbox{Q-learning}, a learning technique originating from RL, to find the optimal transmission policy. \mbox{Q-learning} relies only on the assumption that the underlying system can be modeled as an MDP, and that after taking action $\xt$ in \TSn, the transmitter observes \Ws[\tindex+1], and the instantaneous reward value \WR[\Ws]{\Ws[\tindex+1]}{\xt}. Notice that, the transmitter does not necessarily know \WR[\Ws]{\Ws[\tindex+1]}{\xt} before taking action \xt, because it does not know the next state \Ws[\tindex+1] in advance. In our problem, the immediate reward is  the size of the transmitted packet \Ds; hence, it is readily known at the transmitter.

Eqn. (\ref{eq:Qfun}) indicates that $Q^{\pi}(\Ws,x_i)$ of the current state-action pair can be represented in terms of the expected immediate
reward of the current state-action pair and the state-value function $V^{\pi}(\Ws[\tindex+1])$ of the next state. Note that $Q^{\pi^*}(\Welement,x_i)$  contains all the long term consequences of taking action $x_i$ in state \Welement when following policy $\pi^*$. Thus, one can take the optimal actions by looking only at $Q^{\pi^*}(\Welement,x_i)$ and choosing the action that will yield the highest expected reward (greedy policy).  As a consequence, by only knowing  $Q^{\pi^*}(\Welement,x_i)$, one can derive the optimal policy $\pi^*$ without knowing  \Wtr{\Welement[k]}{x_i} or  \WR{\Welement[k]}{x_i}. Based on this relation, the \mbox{Q-learning} algorithm finds the optimal policy by estimating $Q^{\pi^*}(\Welement,x_i)$ in a recursive manner.
In the $\tindex$th learning iteration $Q^{\pi^*}(\Welement,x_i)$ is estimated by $Q_\tindex(\Welement,x_i)$, which is done by weighting the previous estimate $Q_{\tindex-1}(\Welement,x_i)$  and the estimated expected value of the best action of the next state \Ws[\tindex+1]. In each \TS, the algorithm

\begin{itemize}
  \item observes the current state $\Ws=\Welement\in \Wset$,
  \item selects and performs an action $\xt= x_i\in\mathcal{A}$,
  \item observes the next state $\Ws[\tindex+1]=\Welement[k]\in \Wset$ and the immediate reward \WR{\Welement[k]}{x_i},
  \item updates its estimate of $Q^{\pi^*}(\Welement,x_i)$ using
\begin{equation} \label{eq:QL}
\begin{array}{lcl}
Q_\tindex(\Welement,x_i)&=&(1-\alpha_\tindex)Q_{\tindex-1}(\Welement,x_i)+ \alpha_\tindex \bigl[\WR{\Welement[k]}{x_i}\\
& & +\gamma\max_{x_j\in\mathcal{A}} Q_{\tindex-1}(\Welement[k],x_j)\bigr],
\end{array}
\end{equation}

\end{itemize}
where  $\alpha_\tindex$ is the learning rate factor in the $\tindex$th learning iteration.
If all actions are selected  and performed with non-zero probability, $0\leq\gamma<1$, and the sequence $\alpha_\tindex$ fulfills certain constraints\footnote{The constraints on the learning rate  follow from well-known results in stochastic approximation theory. Denote by $\alpha_{n^k(\Welement,x_i)}$ the learning rate $\alpha_n$ corresponding to the $k$th time action $x_i$ is selected in state \Welement. The constraints on $\alpha_n$ are  $0<\alpha_{n^k(\Welement,x_i)}<1$,  $\sum_{k=0}^{\infty}\alpha_{n^k(\Welement,x_i)}=\infty$, and $\sum_{k=0}^{\infty}\alpha_{n^k(\Welement,x_i)}^2<\infty$,  $\forall \Welement \in \Wset$ and $\forall x_i \in \mathcal{A}$. The second condition is required to guarantee that the algorithm's steps  are large enough to overcome any initial condition. The third condition guarantees that the steps become small enough to assure convergence. Although the use of sequences $\alpha_n$ that meet these conditions assures convergence in the limit, they are rarely used in practical applications.}, the sequence $Q_\tindex(\Welement,x_i)$ is proven to converge to $Q^{\pi^*}(\Welement,x_i)$ with probability $1$ as $\tindex\rightarrow \infty$ \cite{rl:watkins_delayed_rewards}.

With $Q_\tindex(\Welement,x_i)$ at hand the transmitter has to decide for a transmission policy to follow. We recall that, if $Q^{\pi^*}(\Welement,x_i)$ is perfectly estimated by $Q_\tindex(\Welement,x_i)$, the optimal policy is the greedy policy. However, until $Q^{\pi^*}(\Welement,x_i)$ is accurately estimated the greedy policy based on $Q_\tindex(\Welement,x_i)$ is not optimal. In order to estimate  $Q^{\pi^*}(\Welement,x_i)$ accurately, the transmitter should balance the exploration of new actions with the exploitation of known actions.  In exploitation the transmitter follows the greedy policy; however, if only exploitation occurs optimal actions might remain unexplored. In exploration the transmitter takes actions randomly with the aim of discovering better policies and enhancing its estimate of  $Q^{\pi^*}(\Welement,x_i)$. The \mbox{$\epsilon$-greedy} action selection method either takes actions randomly (explores) with probability $\epsilon$ or follows the greedy policy (exploits) with probability  $1-\epsilon$ at each \TS, where $0<\epsilon<1$.

The convergence rate of $Q_\tindex(\Welement,x_i)$ to $Q^{\pi^*}(\Welement,x_i)$  depends on the learning rate $\alpha_\tindex$. The convergence rate decreases with the number of actions, states, and the discount factor $\gamma$, and increases with the number of learning iterations, \LT. See~\cite{Even-Dar2003} for a more detailed study of the convergence rate of the \mbox{Q-learning} algorithm.  \mbox{Q-learning} algorithm is given in Algorithm~\ref{alg:QL}. In Section~\ref{sec:results} the performance of \mbox{Q-learning} in our problem setup is evaluated and compared to other approaches.

\begin{algorithm}[t]
\small\caption{\mbox{Q-learning}} \label{alg:QL}
\begin{algorithmic}
   \STATE \textbf{$\mathbf{1.}$ Initialize:}
   \FOR{each $\Welement \in \Wset,$  $x_i \in \mathcal{A}$}
           \STATE initialize $Q(\Welement,x_i)$ arbitrarily
   \ENDFOR
   \STATE set initial time index $\tindex\leftarrow 1$
   \STATE evaluate the starting state $\Welement\leftarrow\Ws[\tindex]$
   \STATE \textbf{$\mathbf{2.}$  Learning:}
   \REPEAT
    \STATE select action \xt following the \mbox{$\epsilon$-greedy} action selection method
    \STATE perform action $x_i \leftarrow \xt$
    \STATE observe the next state $\Welement[k] \leftarrow \Ws[\tindex+1]$
    \STATE receive an immediate cost \WR{\Welement[k]}{x_i}
    \STATE select the action $x_j$ corresponding to the  $\max_{x_j} Q(\Welement[k],x_j)$
    \STATE update the $Q(\Welement,x_i)$ estimate as follows:

        \STATE \begin{center}$Q(\Welement,x_i)\leftarrow (1-\alpha_\tindex)Q(\Welement,x_i)+ \alpha_\tindex[\WR{\Welement[k]}{x_i} +\gamma \max_{x_j} Q(\Welement[k],x_j)]$ \end{center}
   \STATE update the current state $\Welement\leftarrow\Welement[k]$
   \STATE update the time index $\tindex\leftarrow\tindex+1$
   \UNTIL check stopping criteria $\tindex=\LT$
\end{algorithmic}
\end{algorithm}

\section{Offline Optimization}\label{sec:offline}
In this section we consider the problem setting in Section~\ref{sec:system} assuming that all the future data/energy arrivals as well as the channel variations are known non-causally at the transmitter before the transmission starts. Offline optimization is relevant in applications for which the underlying stochastic processes can be estimated accurately in advance at the transmitter. In general the solution of the corresponding offline optimization problem can be considered as an upperbound on the performance of the online and learning theoretic problems.  Offline approach  optimizes the transmission policy over a realization of the MDP for a finite number of \TSD, whereas the learning theoretic and online optimization problems optimize the expected total transmitted data over an infinite horizon. We recall that an MDP realization is a sequence of state transitions of the data, EH and the channel state processes for a finite number of \TSD. Given an MDP realization in the offline optimization approach, we optimize \xt such that the the expected total transmitted data is maximized.
From~(\ref{eq:opt_problemv1}) the offline optimization problem can be written as follows
\begin{subequations} \label{eq:opt_problemvMILP}
\begin{align}
\max_{\mathbf{X}, \mathbf{B}} & ~~\sum_{\tindex=0}^{\NTS} \gamma^\tindex \xt \Ds \label{eq:obj} \\
\text{s.t.} &~~\xt  \Emin \leq \Bs, \label{eq:const0}\\
&~~\Bs[\tindex+1] \leq  \Bs - \xt \Emin+ \Ehs,\label{eq:const1}\\
&~~0\leq \Bs \leq \NB, \label{eq:const2}\\
&~~\xt\in\{0,1\}, ~~\tindex=0,\dots,\NTS, \label{eq:binconst}
\end{align}
\end{subequations}
where $\mathbf{B}= [\Bs[0] \cdots \Bs[\NTS] ]$  and $\mathbf{X}= [\xt[0] \cdots \xt[\NTS] ]$. Note that we have replaced the equality constraint in (\ref{eq:battery_2}) with two inequality constraints, namely (\ref{eq:const1}) and (\ref{eq:const2}). Hence, the problem in~(\ref{eq:opt_problemvMILP}) is a relaxed version of~(\ref{eq:opt_problemv1}). To see that the two problems are indeed equivalent, we need to show that any solution to~(\ref{eq:opt_problemvMILP}) is also a solution to~(\ref{eq:opt_problemv1}). If the optimal solution to~(\ref{eq:opt_problemvMILP}) satisfies~(\ref{eq:const1}) or~(\ref{eq:const2}) with equality, then it is a solution to~(\ref{eq:opt_problemv1}) as well. Assume that $\mathbf{X}, \mathbf{B}$ is an optimal solution to~(\ref{eq:opt_problemvMILP}) and that for some \tindex, \Bs fulfills both of the constraints~(\ref{eq:const1}) and~(\ref{eq:const2}) with strict inequality whereas the other components satisfy at least one constraint with equality. In this case, we can always find a $\Bs^+>\Bs$ such that at least one of the constraints is satisfied with equality. Since $\Bs^+>\Bs$, (\ref{eq:const0}) is not violated and $\mathbf{X}$ remains to be feasible, achieving the same objective value. In this case, $\mathbf{X}$ is feasible and a valid optimal solution to~(\ref{eq:opt_problemv1}) as well, since $\Bs^+$  satisfies~(\ref{eq:battery_2}).

The problem in (\ref{eq:opt_problemvMILP}) is a \emph{mixed integer linear program} (MILP) problem since it has affine objective and constraint functions, while the optimization variable \xt is constrained to be binary.  This problem is known to be NP-hard; however, there are  algorithms combining relaxation tools with smart exhaustive search methods to reduce the solution time. Notice that, if one relaxes the binary constraint on \xt to \mbox{$0\leq \xt \leq 1$,} (\ref{eq:opt_problemvMILP})  becomes a linear program (LP). This corresponds to the problem in which the transmitter does not make binary decisions, and is allowed to transmit smaller portions of the packets. We call the optimization problem in (\ref{eq:opt_problemvMILP}) the \mbox{complete-problem} and its relaxed version the \mbox{LP-relaxation}.  We define $\mathcal{O}=\{0,1\}^N$ as the feasible set for \BBv  in the \mbox{complete-problem}. The optimal value of the \mbox{LP-relaxation} provides an upper bound on the \mbox{complete-problem}. On the other hand, if the value of \BBv in the optimal solution of the \mbox{LP-relaxation} belong to $\mathcal{O}$, it is also an optimal solution to the \mbox{complete-problem}.

Most available MILP solvers employ an LP based \emph{branch-and-bound} (\BAB) algorithm \cite{opt:Atamturk2005}. In exhaustive search one has to evaluate the objective function for each point of the feasible set $\mathcal{O}$. The \BAB algorithm discards some subsets of $\mathcal{O}$ without evaluating the objective function over these subsets.  \BAB works by generating disjunctions; that is, it partitions the feasible set $\mathcal{O}$ of the \mbox{complete-problem} into smaller subsets,  $\mathcal{O}_k$, and explores or discards each subset $\mathcal{O}_k$ recursively.  We denote  the \mbox{$k$th} active subproblem which solve  (\ref{eq:opt_problemvMILP}) with \BBv constrained to the subset $\mathcal{O}_k\subseteq \mathcal{O}$ by \MILP, and its associated upperbound by $I_k$. The optimal value of \MILP is a lowerbound on the optimal value of the \mbox{complete-problem}. The algorithm maintains a list \List of active subproblems over all the active subsets $\mathcal{O}_k$ created. The feasible solution among all explored subproblems with the highest optimal value  is called the \emph{incumbent}, and its optimal value  is denoted by $I_{max}$. At each algorithm iteration an active subproblem \MILP is chosen, deleted from \List, and its \mbox{LP-relaxation} is solved. Let $\hat{\BBv}^k$ be the optimal $\BBv$ value corresponding to the solution of the \mbox{LP-relaxation} of \MILP, and $I_k^{LP}$ be its optimal value. There are three possibilities: 1) If $\hat{\BBv}^k\in\mathcal{O}_k$, \MILP and its \mbox{LP-relaxation} have the same solution. We update $I_{max}=\max\{I_k^{LP},I_{max}\}$, and all subproblems \MILP[m] in \List for which $I_m\leq I_{max}$ are discarded; 2) If $\hat{\BBv}^k\notin\mathcal{O}_k$ and $I_k^{LP} \leq I_{max}$, then the optimal solution of \MILP can not improve $I_{max}$, and the subproblem \MILP is discarded, and 3) If $\hat{\BBv}^k\notin\mathcal{O}_k$ and $I_k^{LP} > I_{max}$, then \MILP  requires further exploration, which is done by branching it further, i.e., creating two new subproblems from \MILP by branching its feasible set $\mathcal{O}_k$ into two.

For the binary case that we are interested in, a branching step is as follows. Assume that for some $n$, the $\tindex$th  element of $\hat{\BBv}^k$ is not binary, then we can formulate a logical disjunction for the $\tindex$th element of the optimal solution by letting $\BBvar= 0$,  or $\BBvar = 1$.
 With this logical disjunction the algorithm creates two new subsets $\mathcal{O}_{k'}=\mathcal{O}_{k}\cap \{\BBv:\BBvar=1\}$ and $\mathcal{O}_{k''}=\mathcal{O}_{k}\cap \{\BBv:\BBvar=0\}$, which partition $\mathcal{O}_k$ into two mutually exclusive subsets. Note that $\mathcal{O}_{k'}\cup\mathcal{O}_{k''}=\mathcal{O}_k$. The two subproblems, \MILP[k'] and \MILP[k''], associated with the new subsets $\mathcal{O}_{k'}$ and $\mathcal{O}_{k''}$, respectively; are added to \List. The upperbounds $I_{k'}$ and $I_{k''}$ associated  to \MILP[k'] and \MILP[k''], respectively,  are set equal to $I_k^{LP}$.

 After updating \List and $I_{max}$ the \BAB algorithm selects another subproblem \MILP[m] in \List to explore. The largest upperbound associated with the active subproblems in \List is an upperbound on the \mbox{complete-problem}. The \BAB algorithm terminates when \List is empty, in which case this upperbound is equal to the value of the incumbent.  The \BAB algorithm is given in Algorithm~\ref{alg:BB}. In principle, the worst-case complexity of \BAB is $O\left(2^{N}\right)$, same as exhaustive search; however, the average complexity of \BAB is usually much lower, and is polynomial under certain conditions~\cite{Zhang1996}.

\begin{rem}
Notice that, unlike the online and learning theoretic optimization problems, the offline optimization approach is not restricted to the case where $0\leq\gamma<1$. Hence, both the \BAB algorithm and the LP relaxation can be applied to the \TP in~(\ref{eq:opt_problemv3}).
\end{rem}
\begin{algorithm}[t]
\small\caption{\BAB} \label{alg:BB}
\begin{algorithmic}
    \STATE \textbf{$\mathbf{1.}$ Initialize:}
        \STATE $I_{max}=0$, $\mathcal{O}_0=\mathcal{O}$, and $ I_0=\infty$
        \STATE set $\MILP[0]\leftarrow \{\text{solve (\ref{eq:opt_problemvMILP}) s.t. } \BBv \in \mathcal{O}_0\}$
        \STATE $\List\leftarrow\MILP[0]$

    \STATE \textbf{$\mathbf{2.}$ Terminate:}
        \IF{$\List=\emptyset$}
            \STATE $\hat{\BBv}_{max}$ is the optimal solution and $I_{max}$ the optimal value
        \ENDIF

    \STATE \textbf{$\mathbf{3.}$ Select:}
        \STATE choose and delete a subproblem \MILP form \List

    \STATE \textbf{$\mathbf{4.}$ Evaluate:}
        \STATE solve LP-relaxation of \MILP
        \IF{LP-relaxation is infeasible}
            \STATE  go to Step $\mathbf{2}$
        \ELSE
            \STATE let $I_{k}^{LP}$  be its optimal value and $\hat{\BBv}^{k}$ the optimal $\mathbf{X}$ value
        \ENDIF

    \STATE \textbf{$\mathbf{5.}$ Prune:}
        \IF{$I_{k}^{LP} \leq I_{max}$}
            \STATE  go to Step $\mathbf{2}$
        \ELSIF{$\hat{\BBv}^{k}\notin\mathcal{O}_k$}
            \STATE go to Step $\mathbf{6}$
        \ELSE
            \STATE $I_{max}\leftarrow I_k^{LP}$ and $\hat{\BBv}_{max}\leftarrow\hat{\BBv}^{k}$
            \STATE delete all subproblems \MILP[m] in \List with $I_m \leq I_{max}$
        \ENDIF
  \STATE \textbf{$\mathbf{6.}$ Branch:}
        \STATE choose $n$, such that $\hat{\BBv}_n^k$ is not binary
        \STATE set $I_{k'}, I_{k''}\leftarrow I_{k}^{LP}$, $\mathcal{O}_{k'}\leftarrow\mathcal{O}_{k}\cap \{\BBv:\BBvar=1\}$ and $\mathcal{O}_{k''}\leftarrow\mathcal{O}_{k}\cap \{\BBv:\BBvar=0\}$
        \STATE set $\MILP[k']\leftarrow \{\text{solve (\ref{eq:opt_problemvMILP}) s.t. } \BBv \in \mathcal{O}_{k'}\}$ and $\MILP[k'']\leftarrow \{\text{solve (\ref{eq:opt_problemvMILP}) s.t. } \BBv \in \mathcal{O}_{k''}\}$
        \STATE add \MILP[k'] and \MILP[k'']to \List
        \STATE go to Step $\mathbf{3}$

\end{algorithmic}
\end{algorithm}

\section{Numerical Results}\label{sec:results}

To compare the performance of the three approaches that we have proposed, we focus on a sample scenario of the EH communication system presented in Section~\ref{sec:system}. We are interested in comparing the expected performance of the proposed solutions.  For the online optimization approach it is possible  to evaluate the expected performance of the optimal policy $\pi^*$, found using the DP algorithm, by solving (\ref{eq:Vfun}), or evaluating (\ref{eq:Vest}) and averaging over all possible  starting states $\Ws[0]\in\Wset$. In theory, the learning theoretic approach will achieve the same performance as the online optimization approach as the learning time goes to infinity (for $0\leq\gamma<1$); however, in practice the transmitter can learn only for a finite number of \TSD and the transmission policy it arrives at depends on the specific realization of the MDP.  The offline optimization approach optimizes over a realization of the MDP. To find the expected performance of the offline optimization approach  one has to average over infinite realizations of the MDP for an infinite number of \TSD. We can average the performance over only a finite number of MDP realizations and finite number of \TSD. Hence, we treat the performances of the proposed algorithms as a random variable, and use the sample mean to estimate their expected values. Accordingly, to provide a measure of accuracy for our estimators, we also compute the confidence intervals. The details of the confidence interval computations are relegated to the Appendix.

 In our numerical analysis we use parameters based on an \mbox{IEEE802.15.4e} \cite{ieeeWPAN4e} communication system. We consider a \TS length of \mbox{$\Delta_{\TS}=10$ms}, a transmission time of \mbox{$\Delta_{Tx}=5$ ms}, and an available bandwidth of \mbox{$W=2$ MHz}. The fundamental energy unit is \mbox{$2.5~\mu$J} which may account for a vibration or piezoelectric harvesting device \cite{Chalasani2008}, and we assume that the transmitter at each \TS either harvests two units of energy or does not harvest any, i.e., $\Ehset=\{0,2\}$. We denote the probability of harvesting two energy units in \TSn given that the same amount was harvested in \TSn[\tindex-1] by $\phu$, i.e., $\phu\triangleq\Ehtr[2]{2}$.  We will study the effect of \phu and \NB on the system performance and the convergence behavior of the learning algorithm. We set $\Ehtr[0]{0}$, the probability of not harvesting any energy in \TSn when no energy was harvested in \TSn[\tindex-1], to $0.9$. The battery capacity \NB is varied from $5$ to $9$ energy units. The possible packet sizes are $\Ds\in\Dset=\{300,600\}$ bits with state transition probabilities $\Dtr[d_1]{d_1}=\Dtr[d_2]{d_2}=0.9$. Let the channel state at \TSn be $\Cs\in\Cset=\{1.655\cdot 10^{-13},3.311\cdot 10^{-13}\}$ which are two realizations of the indoor channel model for urban scenarios in \cite{doc:ana_cog_doc} with $d=d_{indoor}=55$, $w=3$, $WP_{in}=5$, and $5$~dBm standard deviation, where $d$ is the distance in meters, $w$ the number of walls, and $WP_{in}$ the wall penetration losses. The state transition probability function is characterized by $\Ctr[h_1]{h_1}=\Ctr[h_2]{h_2}=0.9$.

 To find the required energy to reliably transmit a data packet over the channel we consider Shannon's capacity formula for Gaussian channels. The transmitted data in \TSn  is
  \begin{equation}
  \Ds=W\Delta_{Tx}\log_2 \left(1+\frac{\Cs P}{WN_0}\right),
  \end{equation}
  where $P$ is the transmit power and $N_0=10^{-20.4}$ (W/Hz) is the noise power density. In low power regime, which is of special practical interest in the case of machine-to-machine communications or wireless sensor networks with EH devices, the capacity formula can be approximated by $\Ds \simeq \frac{\Delta_{Tx}\Cs P}{\log(2)N_0} $, where $\Delta_{Tx} P$ is the energy expended for transmission in \TSn. Then, the minimum energy required for transmitting a packet  \Ds is given by $\Emin=\fDE{\Cs}=\frac{\Ds\log(2) N_0}{\Cs}$. In general, we assume that the transmit energy for each packet at each channel state is an integer multiple of the energy unit. In our special case, this condition is satisfied as we have $\mathcal{E}_u=\{1,2,4\}$, which correspond to transmit power values of $0.5, 1$ and \mbox{$2$ mW}, respectively. Numerical results for the \DP, in which the transmitter might terminate its operation at each \TS with probability $\gamma$ is given in Section~\ref{sec:DDSP} whereas the \TP is examined in Section~\ref{sec:TP}.

\subsection{\DP}\label{sec:DDSP}
We generate $T=2000$ realizations of $N=100$ random state transitions and examine the performance of the proposed algorithms for $\gamma=0.9$. In particular, we consider the \mbox{LP-relaxation} of the offline optimization problem, the offline optimization problem with the \BAB algorithm\footnote{Reference~\cite{opt:Atamturk2005} presents a survey on software tools for MILP problems. In this paper we use the \BAB toolbox provided in~\cite{sw:lpsolve}. In particular, \BAB is set up with a $20$ seconds timeout. For the particular setting of this paper, the \BAB algorithm found an optimal solution, within the given timeout, $99.7\%$ of the times.}, the online optimization problem with PI, the learning theoretic approach with \mbox{Q-learning}\footnote{We use the $\epsilon$-greedy action selection mechanism with $\epsilon=0.07$, and set the learning rate to $\alpha=0.5$.}. We have considered a greedy algorithm which assumes a causal knowledge of \Bs, \Ds and \Cs, and transmits a packet whenever there is enough energy in the battery ignoring the Markovity of the underlying processes.

Notice that the \mbox{LP-relaxation} solution is an upper bound on the performance of the \Milp optimization problem, which, in turn, is an upper bound on the \PI problem. At the same time the performance of the  \PI optimization problem is an upper bound on the \QL and the \GP approaches.

\begin{figure}[tb!]
  \centering
  \includegraphics[width=0.5\textwidth]{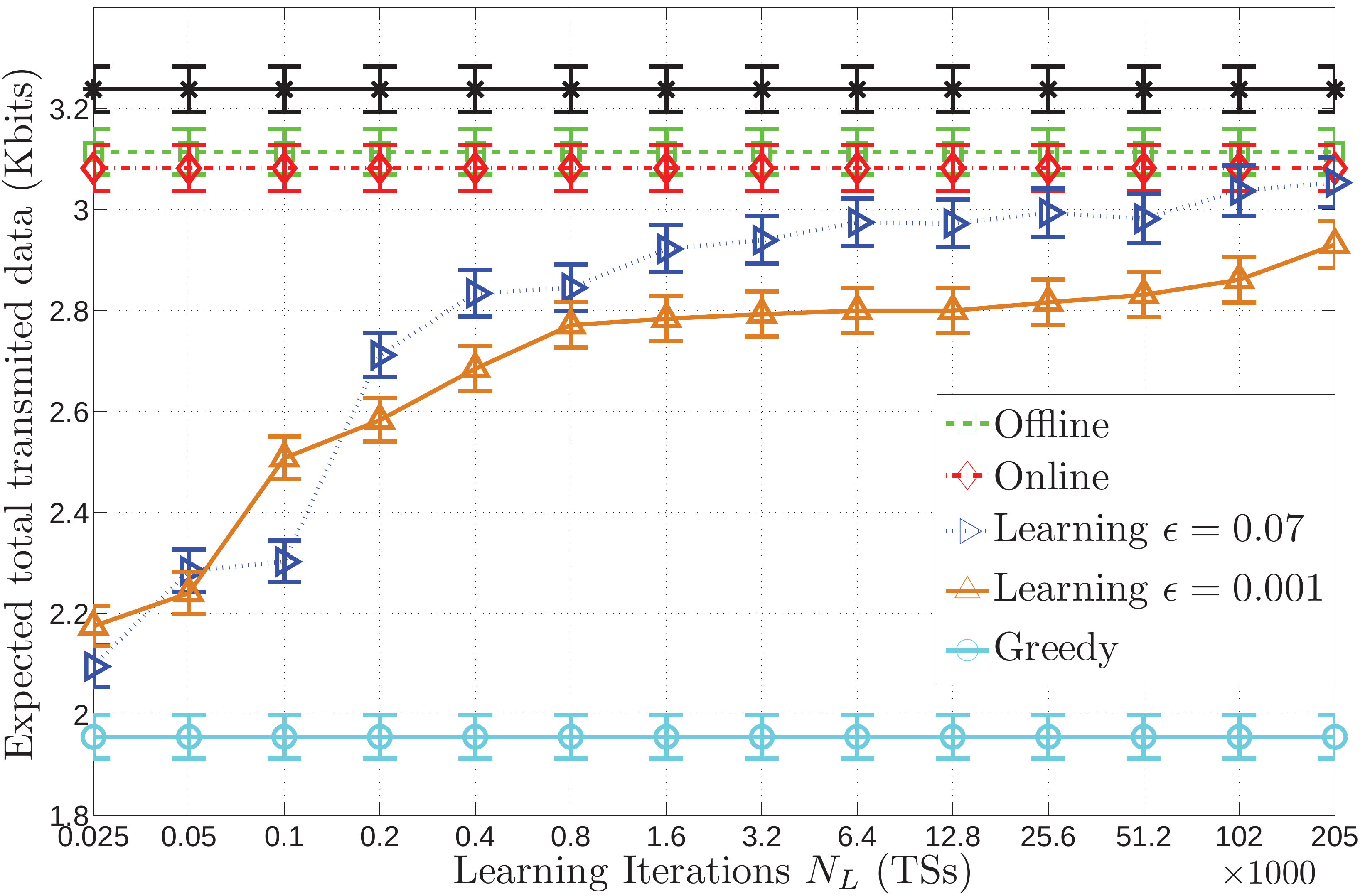}\\
  \vspace{-0.5cm}
  \caption{Expected total transmitted data with respect to the learning time \LT, with $\phu=0.9$, and \NB=5.}\label{fig:time}\vspace{-0.5cm}
\end{figure}

In Figure~\ref{fig:time} we illustrate, together with the performance of the other approaches, the expected total transmitted data by the \QL approach against the number of learning iterations, \LT. We can see that for \mbox{$\LT>200$ \TSD}  the learning theoretic approach ($\epsilon=0.07$) reaches  $85\%$ of the performance achieved by online optimization, while for \mbox{$\LT> 2\cdot 10^5$ \TSD} it reaches $99\%$.  We can conclude that the \QL approach is able to learn the optimal policy with increasing accuracy as \LT increases.  Moreover, we have investigated the exploration/exploitation tradeoff of the learning algorithm, and we have observed that for low exploration values ($\epsilon=0.001$) the learning rate decreases, compared to moderate exploration values ($\epsilon=0.07$). We also observe  from Figure~\ref{fig:time} that the performance of the \GP algorithm is notably inferior compared to the other approaches.

\begin{figure}[t]
     \begin{center}
        \subfigure[$\phu=\{0.5,\dots,0.9\}$ and  $\NB=5$.]{%
           \label{fig:energy}
\includegraphics[width=0.5\textwidth]{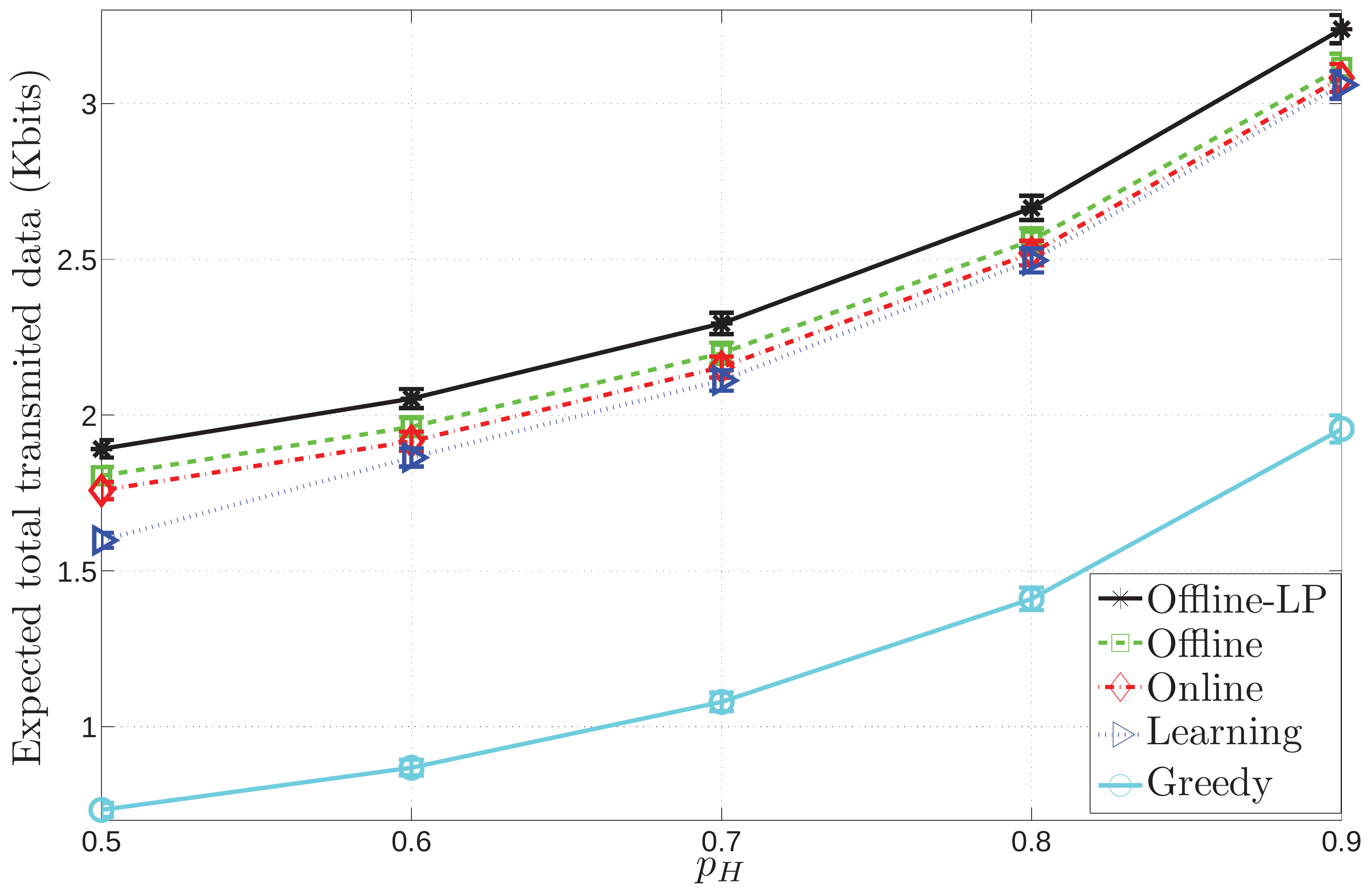}
        }%
        \\
       \subfigure[$B_{max}=\{5,\ldots,9\}$ and  $\phu=0.9$.]{%
           \label{fig:rvsb}
\includegraphics[width=0.5\textwidth]{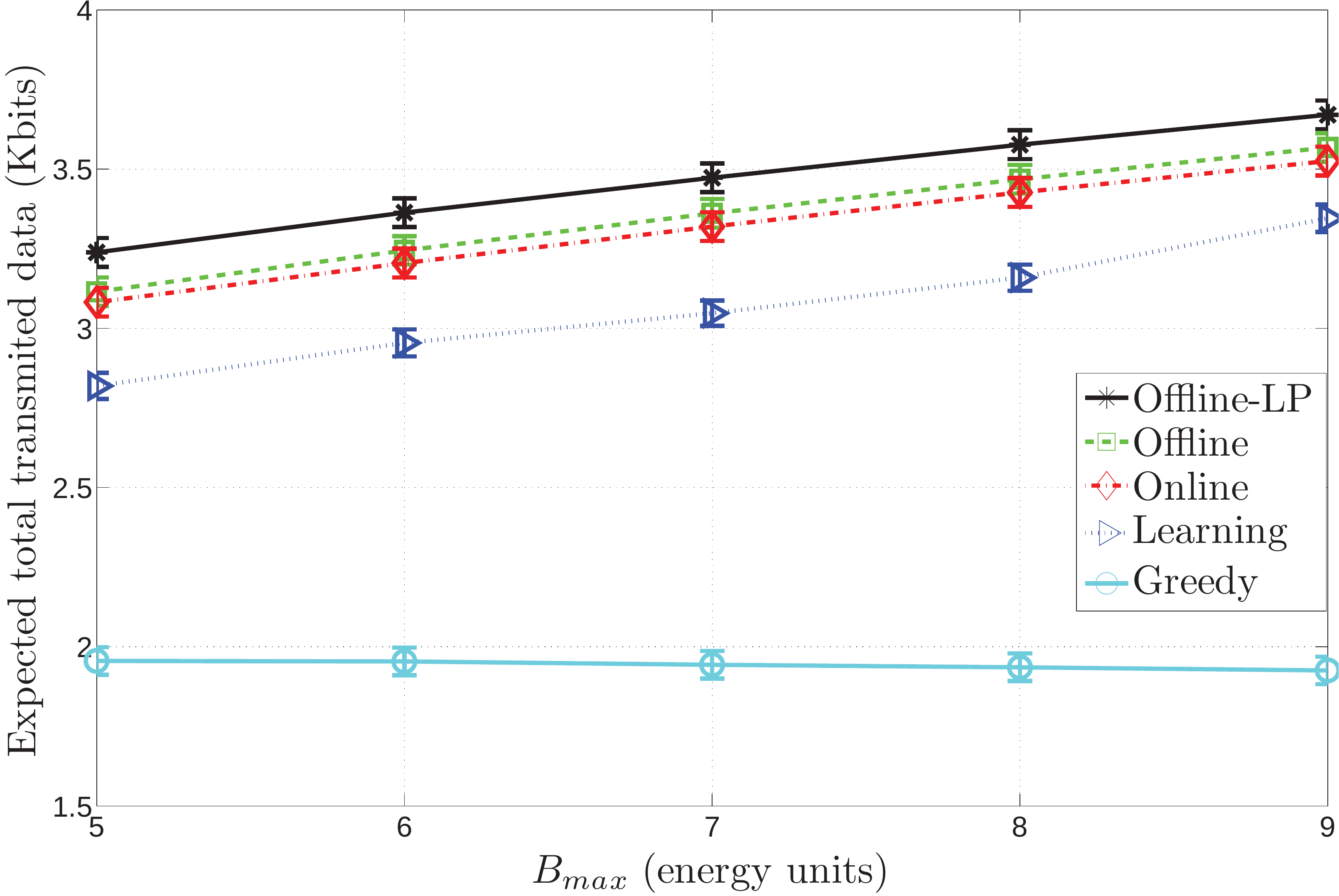}
        }\\
    \end{center}

    \caption{Performance comparison of proposed algorithms for the \DP.}
      \vspace{-0.5cm}
   \label{fig:subfigures}
\end{figure}

Figure~\ref{fig:energy} displays the expected total transmitted data for different \phu values. We consider \mbox{$\LT=10^4$ \TSD} for the learning theoretic approach since short learning times are more practically relevant. As expected, performance of all the approaches increase as the average amount of harvested energy increases with \phu. The offline approach achieves, on average, $96\%$ of the performance of the offline-LP solution. We observe that the \QL approach converges to the online optimization performance with increasing \phu, namely its performance is $90\%$ and $99\%$ of that of the online approach for $\phu=0.5$ and $\phu=0.9$, respectively. It can also be seen that the \PI optimization achieves $97\%$ of the performance of the \Milp optimization when $\phu=0.5$, while for $\phu=0.9$ it reaches $99\%$. This is due to the fact that the underlying EH process becomes less random as \phu increases; and hence, the online algorithm can better estimate its future states and adapt to it. Additionally, we observe from Figure~\ref{fig:energy} that the performance of the \GP approach reaches a mere $60\%$ of the \Milp optimization.

%

In Figure~\ref{fig:rvsb} we show the effect of the battery size, $B_{max}$, on the expected total transmitted data for \mbox{$\LT=10^4$ \TSD}. We see that the expected total transmitted data increases with $B_{max}$ for all the proposed algorithms but the greedy approach. Overall, we observe that the performance of the online optimization is approximately  $99\%$ that of the offline optimization. Additionally, we see that the learning theoretic approach reaches at least $91\%$  of the performance of the online optimization.  Although only a small set of numerical results is presented in the paper due to space limitations, we have executed exhaustive numerical simulations with different parameter settings and observed similar results.

\subsection{\TP}\label{sec:TP}

In the online and learning theoretic formulations, the \TP in (\ref{eq:opt_problemv3})  falls into the category of average reward maximization problems, which cannot be solved with \mbox{Q-learning} unless a finite number of \TSD is specified, or the MDP presents absorbing states. Alternatively, one can take advantage of the average reward RL algorithms. Nevertheless, the convergence properties of these methods are not yet well understood. In this paper we consider \mbox{R-learning}\footnote{In R-learning \WR{\Welement[k]}{x_i} in (\ref{eq:QL}) is substituted by $\WRAdj{\Welement[k]}{x_i}= \WR{\Welement[k]}{x_i}-\rho_\tindex$, where $\rho_{\tindex} = (1-\beta)\rho_{\tindex-1}+\beta \bigl [\WR{\Welement[k]}{x_i} +\max_{x_j\in\mathcal{A}} Q_{\tindex-1}(\Welement[k],x_j) -\max_{x_j\in\mathcal{A}} Q_{\tindex-1}(\Welement[j],x_j) \bigr ]$, $0\leq\beta\leq1$, and $\rho_\tindex$ is  updated in \TSn only if a non-exploratory action is taken.}~\cite{rl:Mahadevan1996} which is similar to \mbox{Q-learning}, but is not proven to converge.

Similarly, for the online optimization problem, the policy evaluation step in the PI algorithm is not guaranteed to converge for $\gamma=1$. Instead, we use relative value iteration (RVI)~\cite{dp:Putterman2005}, which is a DP algorithm, to find the optimal policy in average reward MDP problems.

 In our numerical analysis for the \TP, we consider the \mbox{LP-relaxation} of the offline optimization problem, the offline optimization problem with the \BAB algorithm, the online optimization problem with RVI, the learning theoretic approach with R-learning\footnote{We use the same action selection method as the \mbox{Q-learning} algorithm in Section~\ref{sec:DDSP}.}, and finally,  the greedy algorithm. For evaluation purposes we average over $T=2000$ realizations of $N=100$ random state transitions.

\begin{figure}[t]
     \begin{center}
        \subfigure[Average throughput versus \LT for $\phu=0.9$]{%
           \label{fig:time_av}
\includegraphics[width=0.51\textwidth]{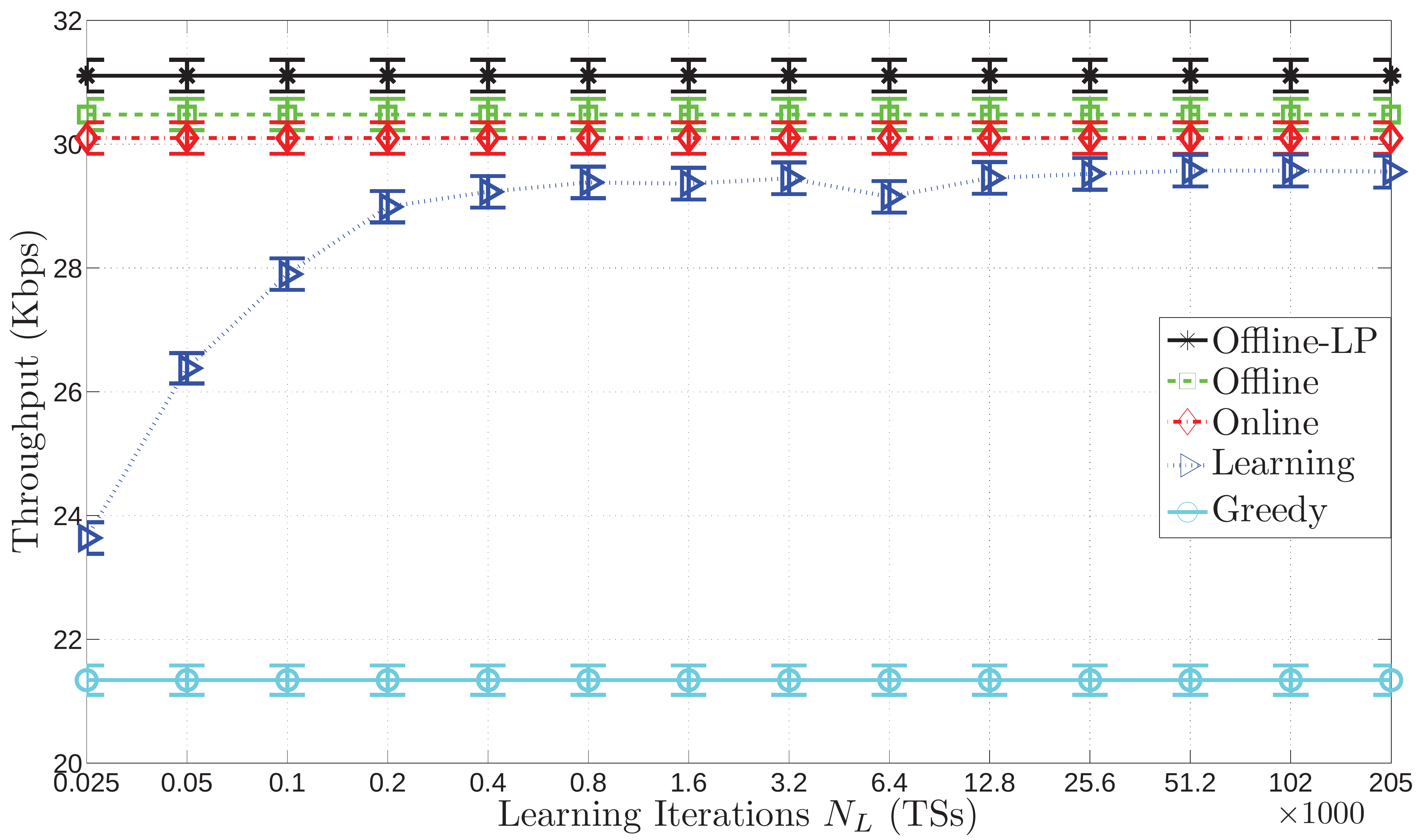}
        }%
        \\
        \subfigure[Average throughput for $\phu=\{0.5,\dots,0.9\}$.]{%
           \label{fig:energy_av}
\includegraphics[width=0.51\textwidth]{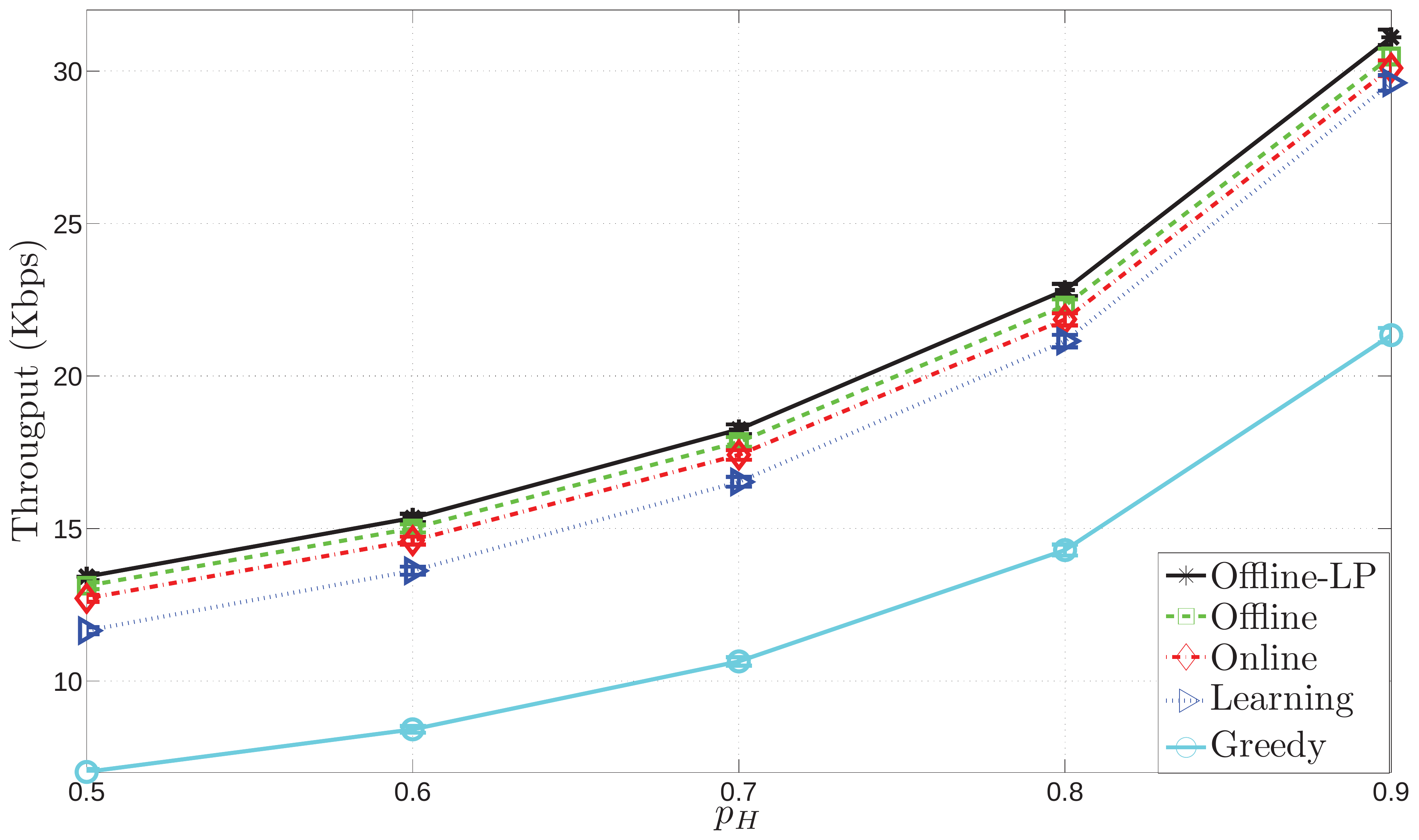}
        }\\
    \end{center}

    \caption{Performance comparison of the proposed algorithms for the \TP for $\NB=5$.}
\vspace{-1cm}
   \label{fig:subfigures}
\end{figure}

In Figure~\ref{fig:time_av} we illustrate, together with the performance of the other approaches, the throughput achieved by the \QL approach against the number of learning iterations, \LT. We observe that for \mbox{$\LT > 200$ \TSD}  the learning algorithm reaches $95\%$ of the performance achieved by online optimization, while for \mbox{$\LT >2\cdot10^5$ \TSD} the performance is $98\%$ of the performance of the \PI optimization approach. Notably the \QL approach performance increases with \LT; however, in this case the performance does not converge to the performance of the \PI optimization approach. As before the \GP algorithm is notably inferior compared to the other approaches.


Figure~\ref{fig:energy_av} displays the throughput for different \phu values. We plot the performance of the \QL approach for \mbox{$\LT=10^4$ \TSD} and $\epsilon=0.07$. As expected, performance of all the approaches increase as the average amount of harvested energy increases with \phu. It can be seen that the \PI approach achieves, on average, $95\%$ of the performance of the \Milp approach. This is in line with our finding in Figure~\ref{fig:energy}. The throughput achieved by the \QL approach achieves $91\%$ of the \PI optimization throughput for $\phu=0.5$ and $98\%$ for $\phu=0.9$. Similarly to Figure~\ref{fig:energy}, the learning theoretic and the online optimization performances, compared to that of the offline optimization, increase when the underlying Markov processes are less random. Similarly to the \DP, the greedy algorithm shows a performance well below the others. We observe that, although the convergence properties of the \mbox{R-learning} are not well understood it has a similar behavior to \mbox{Q-learning}, in practice.


\section{Conclusions}\label{sec:conclusions}

We have considered a point-to-point communication system in which the transmitter has an energy harvester and a rechargeable battery with limited capacity. We have studied optimal communication schemes under strict deadline constraints. Our model includes stochastic data/energy arrivals and a time-varying channel, all modeled as Markov processes. We have studied the \DP, which maximizes the expected total transmitted data during the transmitter's activation time. Considering various assumptions regarding the information available at the transmitter about the underlying stochastic processes; online, learning theoretic and offline optimization approaches have been studied. For the learning theoretic and the online optimization problems the communication system is modeled as an MDP, and the corresponding optimal transmission policies have been identified. A \mbox{Q-learning} algorithm has been proposed for the learning theoretic approach, and as the learning time goes to infinity its performance has been shown to reach the optimal performance of the online optimization problem, which is solved here using policy iteration algorithm. The offline optimization problem has been characterized as a mixed integer linear program problem, and its optimal solution through the branch-and-bound as well as a linear program relaxation have been presented.

Our numerical results have illustrated the relevance of the learning theoretic approach for practical scenarios. For practically relevant system parameters, it has been shown that, the learning theoretic approach reaches $90\%$ of the performance of the online optimization after a reasonable small number of learning iterations. Accordingly, we have shown that smart and energy-aware transmission policies can raise the performance from $60\%$ up to $90\%$ of the performance of the offline optimization approach compared to the greedy transmission policy. We have also addressed the \TP and made similar observations despite the lack of theoretical convergence results.

\appendix \label{app1}
In the discounted sum data problem we are interested in estimating $\bar{X}=\expected{\lim_{\NTS\rightarrow\infty} \sum_{n=0}^{\NTS}\gamma^n\xt\Ds}{}$, where \xt is the action taken by the transmitter which is computed using either the offline optimization, online optimization or the learning theoretic approach, and \Ds is the packet size in the $\tindex$th \TS. An upper bound on $\bar{X}$ can be found as
\begin{equation}\label{eq:upb1}
\bar{X}\leq\underbrace{\expected{ \sum_{n=0}^{N}\gamma^n\xt\Ds}{}}_{\bar{X}_N} + \underbrace{\Ds[\text{max}]\frac{\gamma^N}{1-\gamma}}_{\epsilon_N},
\end{equation}
which follows by assuming that after \TSn[N] all packets arriving at the transmitter are of size $\Ds[\text{max}] \geq \Delement$ for all $\Delement \in \Dset$, that there is enough energy to transmit all the arriving packets, and that, $0\leq\gamma<1$. Notice that the error $\epsilon_N$ decreases as an exponential function of $N$. Then $\bar{X}$ is constrained by
\begin{equation}
\bar{X}_N  \leq \bar{X} \leq \bar{X}_N + \epsilon_N.
\end{equation}
Now that we have gauged the error $\epsilon_N$ due to not considering an infinite number of \TSD in each MDP realization, we consider next the error due to estimating $\bar{X}_N$ over a finite number of MDP realizations. We can rewrite $\bar{X}_N$ as
\begin{equation}
\bar{X}_N  = \lim_{T \rightarrow \infty}\frac{1}{T} \sum_{t=0}^{T} \left ( \sum_{n=0}^{N} \gamma^n\xt^t\Ds^t  \right),
\end{equation}
where $\xt^t$ and $\Ds^t$ correspond to the action taken and data size in the \TSn of the $t$th MDP realization, respectively. We denote by $\hat{X}_N^T$  the sample mean estimate of $\bar{X}_N$ for $T$ realizations as:
\begin{equation}
\hat{X}_N^T  = \frac{1}{T} \sum_{t=0}^{T} \left ( \sum_{n=0}^{N} \gamma^n\xt^t\Ds^t  \right).
\end{equation}
 Using the Central Limit Theorem, if $T$ is large, we can assume that $\hat{X}_N^T$ is a random variable with normal distribution and by applying the Tchebycheff inequality~\cite{Papoulis1965} we can compute the confidence intervals for $\hat{X}_N^T$
\begin{equation}
P(\hat{X}_N^T-\epsilon_T < \bar{X}_N  < \hat{X}_N^T+\epsilon_T )=\delta,
\end{equation}
where $\epsilon_T \triangleq t_{\frac{1+\delta}{2}}(T) \frac{\hat{\sigma}}{\sqrt{T}}$, with $t_{a}(b)$  denoting the Student$-t$ $a$ percentile for $b$ samples and the variance  $\hat{\sigma}$ is estimated using
\begin{equation}
\hat{\sigma}^2=\frac{1}{T}\sum_{t=0}^T \left(\sum_{n=0}^{N} \xt^t\Ds^t - \hat{X}_N^T \right )^2.
\end{equation}
Finally, the confidence interval for the estimate $\hat{X}_N^T$ of $\bar{X}$ is
\begin{equation}
P(\hat{X}_N^T-\epsilon_T < \bar{X}  < \hat{X}_N^T+\epsilon_T + \epsilon_N)=\delta.
\end{equation}

where $\epsilon_N$ is defined in (\ref{eq:upb1}).
In our numerical analysis we compute the confidence intervals for $\delta=0.9$.
\begin{rem}
In the throughput optimization problem we assume that, given the stationarity of the underlying Markov processes, the expected throughput achieved in a sufficiently large number of \TSD is the same as the expected throughput over an infinite horizon. Thus, by setting $\epsilon_N$ to zero, the computation of the confidence intervals for the \TP is analogous to the \DP.
\end{rem}
\bibliographystyle{IEEEtran}
\bibliography{IEEEabrv,Totabiblio}
\flushend
\end{document}